\documentclass[journal]{IEEEtran}
\ifCLASSINFOpdf
\else
\fi
\usepackage{dblfloatfix}
\usepackage{cite}
\usepackage{hyperref}
\usepackage{graphicx}
\usepackage{amsmath}
\usepackage{subfig}
\usepackage{booktabs}
\usepackage{flushend}
\usepackage{amssymb}
\usepackage{algpseudocode}
\usepackage{algorithm}

\begin{document}
%
\title{Face Manifold: Manifold Learning for Synthetic Face Generation}
%
%
%

\author{Kimia~Dinashi,
        Ramin~Toosi, and
        Mohammad Ali~Akhaee
\thanks{Authors are with the School of Electrical and Computer Engineering, College of Engineering, University of Tehran, Tehran, Iran, email:\{kimiadinashi, r.toosi, akhaee\}@ut.ac.ir}}

\maketitle
\begin{abstract}
	Face is one of the most important things for communication with the world around us. It also forms our identity and expressions. Estimating the face structure is a fundamental task in computer vision with applications in different areas such as face recognition and medical surgeries. Recently, deep learning techniques achieved significant results for 3D face reconstruction from flat images. The main challenge of such techniques is a vital need for large 3D face datasets. Usually, this challenge is handled by synthetic face generation. However, synthetic datasets suffer from the existence of \textit{non-possible} faces. Here, we propose a face manifold learning method for synthetic diverse face dataset generation. First, the face structure is divided into the shape and expression groups. Then, a fully convolutional autoencoder network is exploited to deal with the \textit{non-possible} faces, and, simultaneously, preserving the dataset diversity. Simulation results show that the proposed method is capable of denoising highly corrupted faces. The diversity of the generated dataset is evaluated qualitatively and quantitatively and compared to the existing methods. Experiments show that our manifold learning method outperforms the state of the art methods significantly.
\end{abstract}


%
\IEEEpeerreviewmaketitle

\section{Introduction}

\IEEEPARstart{F}{ace} can indicate different and complex thoughts and feelings through gestures. Alongside with our emotions, the geometric features of our faces form our identity. Recovering the structure of a face is an important task in computer vision with fundamental role in various applications. For example in the face recognition task, a received face should be aligned to form a neutral frontal face to remove the variation of pose in the same subject \cite{hassner2015effective,masi2019learning}. As another example, in medical applications, having the face structure allows better planning for operations and surgeries \cite{sela2016customized,goto2018comparison}.

Recovering the structure of a face from a flat image is a challenging task, since there exist a vast number of degrees of freedom and flexibility. Camera projection, lighting condition, texture, and head position and orientation are the main sources of ambiguity. There exist numerous methods for the face construction using a single image. Some methods, characterize a face with limited number of parameters \cite{sirovich1987low,meytlis2007dimensionality}. These methods are based on the fact that human faces share similar characteristics. A popular example of such methods is 3D morphable model(3DMM) \cite{blanz1999morphable}, which has been used in many applications such as face reconstruction \cite{bas2016fitting,huber2016multiresolution}, recognition \cite{zhu2015high}, digital makeup \cite{scherbaum2011computer}, and synthetic face data generation \cite{dou2017end}. In 3DMM representation of faces, the low dimensional parameter space is derived from a face example dataset, which has a great influence on the resulting model. The main drawback of this representation is that it produces unsatisfactory outputs when the input face deviates largely from the example faces in the dataset. There also exist methods that use RGB-D cameras \cite{bouaziz2013online,Li2013,Hsieh2015}. However, RGB-D cameras such as Microsoft Kinect are not still common and not of high resolution at low cost as RGB cameras are. Shape-from-shading (SFS) \cite{zhang1999shape,Durou2008} is another computer vision technique that is employed in the face reconstruction problem. SFS could potentially result in ambiguous results \cite{Prados2006} and need additional information in the reconstruction phase such as face symmetry \cite{WenYiZhao2002,Zhao2001}, reference face \cite{Kemelmacher-Shlizerman2011}, multi-view images \cite{Wu2011,Ichim2015}, and unconstrained face images \cite{Roth2015,Roth2017}.

With rapid growth and success of deep learning techniques in various computer vision areas, researchers also used this technique in 3D face reconstruction problem \cite{Jourabloo2016,Richardson2017,Tran2017}. An image-to-image translation network which forms a pixel-based mapping from the input image to the depth image is introduced in \cite{Sela2017}. CoarseNet and FineNet are introduced in \cite{Richardson2017} to derive the face structure in a coarse-to-fine fashion. Another coarse-to-fine framework is also proposed in \cite{Guo2018} for monocular videos. Reconstruction under two extreme conditions, i.e. out-of-plane rotation and occlusion is studied in \cite{Tran_2018_CVPR}.
Zhu \textit{et. al.} also employed convolutional neural networks for face alignment purpose \cite{Zhu2019}. Their method could align faces in large poses up to $90^\circ$.

A common point between all mentioned methods which are based on deep learning frameworks is the lack of large labeled dataset. All methods need a large pair of input image and its corresponding true 3D representation. Most of the methods follow pipelines like the one suggested in \cite{Richardson2016}. Each synthetic face is constructed by drawing random vectors for \textit{identity}, \textit{expression}, and \textit{texture} from a face morphable model (e.g. 3DMM). As we mentioned, models like 3DMM are highly dependent on the example dataset. Moreover, the random vectors are drawn from a Gaussian distribution which makes them concentrate around the average face. Another disadvantage of this pipeline is the generation of \textit{non-possible} faces. This is because of the fact that the parameters cover a wide range of positions for different parts of a face in a weekly related manner. However, the different parts of a face are highly related to each other. Thus, a \textit{possible} human face is only exist in a small subspace of this representation, which is called \textit{face manifold} in this paper.

Manifold learning is a process of learning the geometric and topological properties
of samples given that the input data are sampled from a smooth manifold \cite{4359350}.
To the best of our knowledge, there exist no complete research on face manifold in the literature. However, manifold learning is a common technique in some areas like human motion \cite{aristidou2018inverse}, mechanics \cite{ibanez2018manifold}, signal processing \cite{8304694}, and time series analysis \cite{8440718}. By learning a manifold for a human face, most of the non-possible faces resulted by the different synthetic data generators could be corrected or eliminated. As a consequence, the reconstruction methods based on the deep learning techniques, which need a large number of labeled data, could be trained on a large and diverse datasets.

In this paper, we propose a face manifold learning method based on the convolutional autoencoders. First, we divide a face structure into two sets of parameters related to the shape and expression features, in the 3DMM method. Thereafter, a fully convolutional autoencoder is designed to correct the corrupted data. This also could be seen as a noise removing procedure, where the noise is the source of corruption. To train the proposed network, two datasets are used. The input data is a randomly corrupted version of 3DMM parameters and the output is the original \textit{clean} data. Results are evaluated with quantitative and qualitative criteria. The proposed algorithm leads to a promising and high quality results and we can show that by using this method, it is possible to generate a highly diverse synthetic face dataset.

The rest of the paper is organized as follows. In Section \ref{sec:PM}, we go through the details of the proposed method. The evaluation of the proposed method is detailed in Section \ref{sec:exp}. Finally, Section \ref{sec:con} concludes the paper. Code is publicly available on GitHub\footnote{ \url{https://github.com/SCL-UT/face-manifold}.}.

\section{The Proposed Method} \label{sec:PM}
\subsection{Face Model}
3D morphable models are derived from applying principal components analysis (PCA) \cite{jolliffe2003principal} to a set of 3D face scans. Then, a 3D face could be obtained by specifying the coefficients of each of the bases of the face space. As suggested in \cite{chu20143d}, these bases are divided into two sets; first, the bases describing identity of a neutral face, and the second ones describing facial expressions. In a 3DMM model, a face containing $N$ mesh vertices, could be characterized as follows:
\begin{equation}\label{Eq1}
	S = \mu_{S}+\alpha_{id} A_{id}+\alpha_{exp} A_{exp}.
\end{equation}
where $A_{id} \in\rm I\!R^{199\times3N}$ and $A_{exp} \in\rm I\!R^{29\times3N}$ are the identity and expression bases, respectively, and $\mu_S \in\rm I\!R^{1\times3N}$ is the average face. $\alpha_{id} \in\rm I\!R^{1\times199} $ and $\alpha_{exp} \in\rm I\!R^{1\times29}$ are the identity and expression coefficients respectively, which are regarded as the only parameters of this parametric model. The 3DMM bases employed in this work gain from the Basel face model (BFM) \cite{paysan20093d} for face shapes and FaceWarehouse \cite{cao2013facewarehouse} for face expressions. In this model, the final vector $S$, representing a 3D face, contains $53,215$ 3D vertices. Therefore, employing this representation leads to a significant dimensionality reduction from thousands of vertices to only 228 parameters.

\begin{figure}
	\centering
	\includegraphics[width=3.5in]{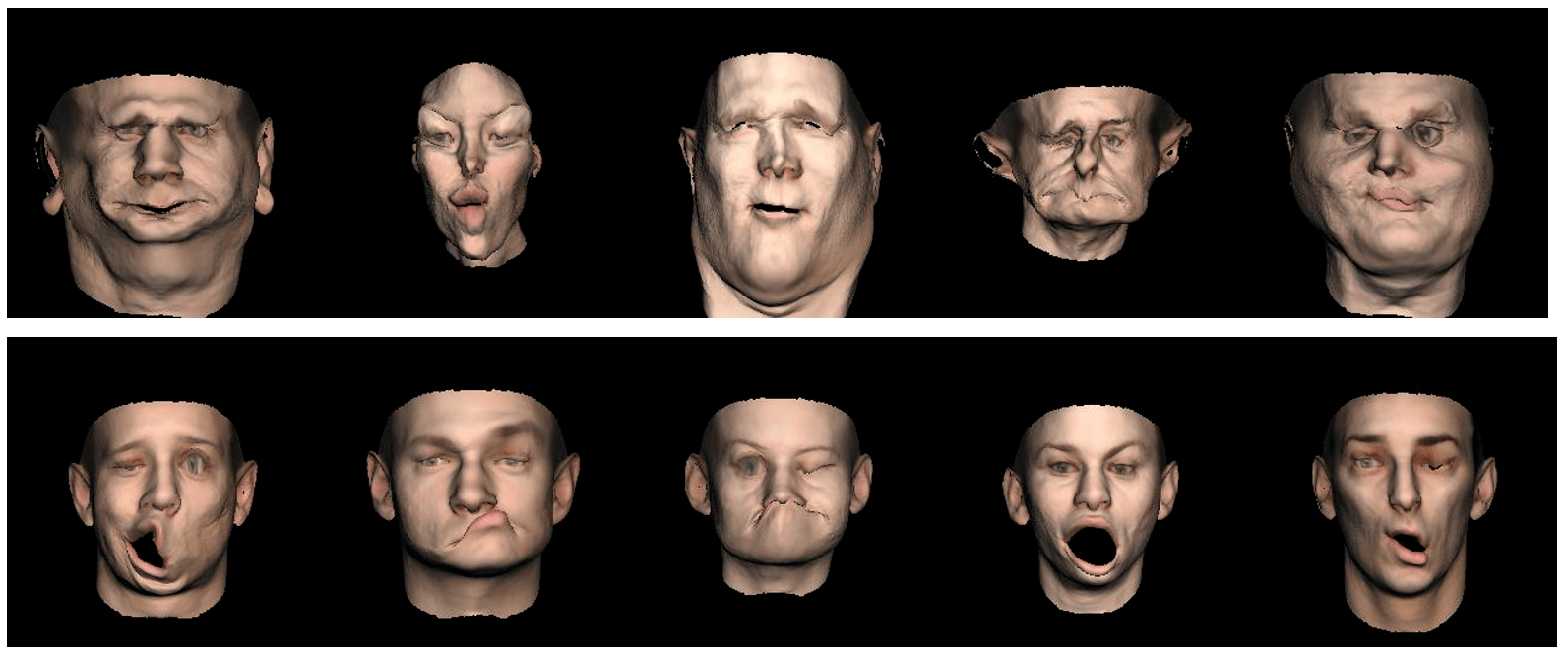}
	\caption{Generated faces by drawing 3DMM parameters from a uniform distribution instead of normal distribution, which have very extreme shapes or expressions. Top row: faces with noisy shapes. Bottom row: faces with noisy expressions.}\label{Fig1}
\end{figure}

As discussed in \cite{blanz1999morphable}, the probability for each parameter follows a Gaussian distribution with the variance equivalent to its corresponding eigenvalue, independently from other parameters. As stated before, many of the existing methods use synthetic datasets for training a network to learn 3D face reconstruction task \cite{Richardson2016, dou2017end, Guo2018, Richardson2017, Sela2017}. In order to generate a synthetic dataset, 3DMM parameters are drawn from the mentioned Gaussian distribution. However this method leads to concentration of generated faces around the mean face and very limited variations in the shape and expression. For instance, in the case of expression parameters, most of the expressions would be similar to the neutral expression. Thus the synthetic dataset would be very restricted far from real face images. Training a neural network on the aforementioned dataset limits the ability of the network in accurate prediction for face images with varying shapes and expressions. As a solution to this problem, one may draw 3DMM parameters from a uniform distribution or within a larger interval instead of the traditional method. This would increase diversity in generated faces, but at the same time decreases their possibility of being a real human face. In other words, the generated faces become more probable to be corrupted or \textit{non-possible}. In Fig. \ref{Fig1} some instances of non-possible faces generated by this method are shown. The interval for each of the parameters is chosen $n$ times of its corresponding eigenvalue. That in the shape and expression parameters $n$ is considered $10$ and $15$, respectively. Through this section, we propose a method to correct this non-possible faces. In other words, correcting their corruption, in such a manner that their deviation from the mean face would be preserved as much as possible. To aim this purpose, two separate convolutional neural networks, one for expression and the other one for shape, are trained in order to learn the human face manifold. Afterwards, the trained networks are used to map non-possible faces to possible ones without losing their particular characteristics in terms of shape and expression.

\subsection{Network Structure}
The mapping from a set of noisy 3DMM parameters to its valid version is learned by a convolutional autoencoder network. An autoencoder structure consists of two parts: i) the encoder part generates a representation of the input with usually reduced dimension by extracting main features while ignoring noise and ii) the decoder part which is responsible for generating a clean version of the input from this compact representation. Autoencoders are considered as effective tools in learning data compression and denoising tasks. The main challenge of this work is in fact  denoising the 3DMM parameters, where an autoencoder network appears to be an appropriate choice. In the following, the structure of the proposed convolutional autoencoder network would be described in details.

\begin{figure*}[!t]
	\centering
	\includegraphics[width=5.5in]{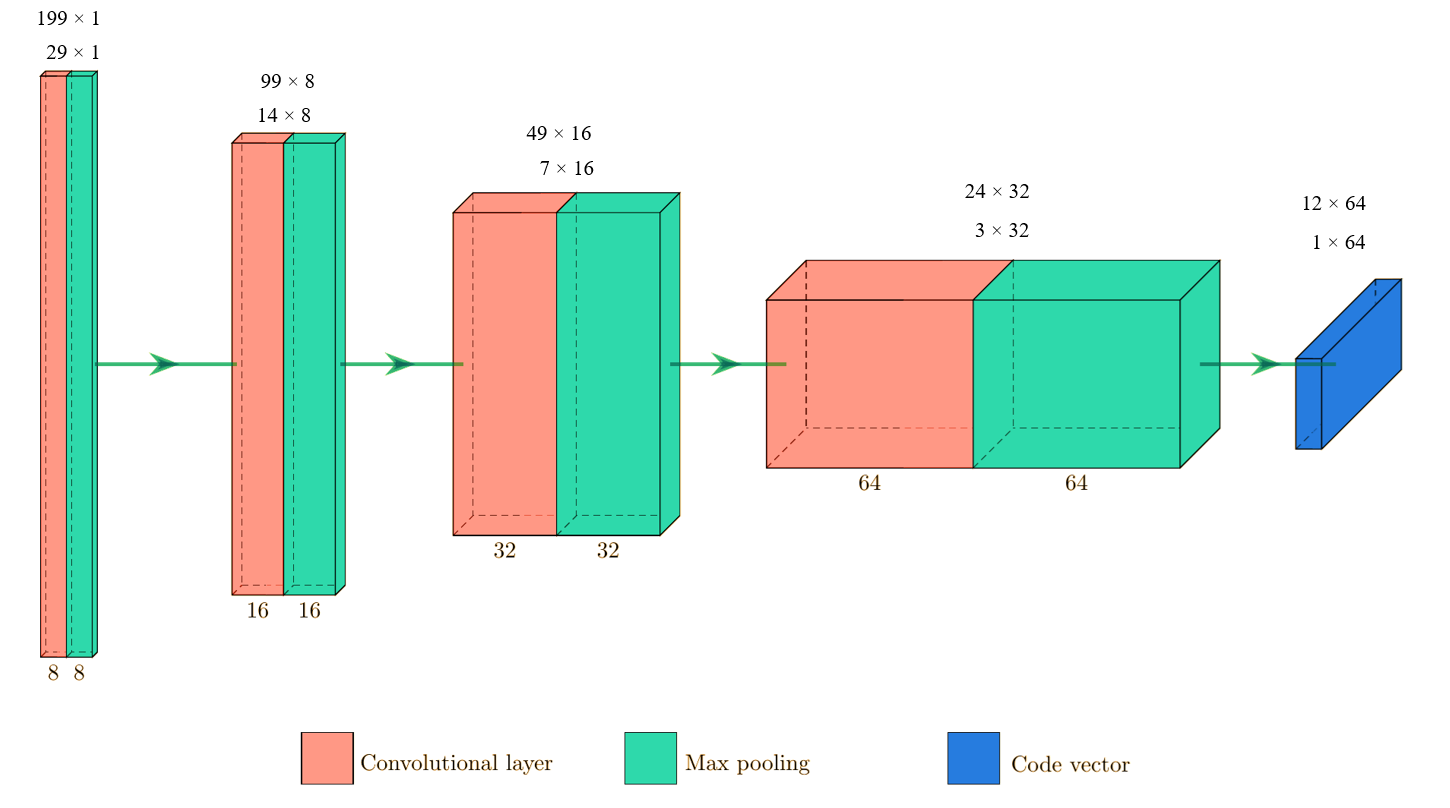}
	\caption{The encoder part of the proposed convolutional autoencoder. Each layer consists of several channels of $3\times1$ convolution filters and $2\times1$ poolings. Number of channels in layers 1, 2, 3, and 4 is respectively 8, 16, 32, and 64. The numbers indicate the dimensions of input feature maps in each layer, the top row for the shape and the bottom row for the expression parameters.}\label{Fig3}
\end{figure*}

Here, an effective yet light-weighted convolutional autoencoder architecture is selected. Empirical experiments demonstrate that considering and training an eight-layer symmetrical autoencoder as both the shape and expression networks, enables promising results in denoising 3DMM parameters. The encoder part of the proposed structure is detailed in Fig. \ref{Fig3}. The architectural symmetry implies that the decoder is the reverse of the encoder, i.e. the order of layers and also the operation of each layer in the encoder is reversed in the decoder. In the encoder, each layer consists of multiple channels of one-dimensional convolution stacked by a max-pooling layer. Alternately, the max-unpooling stacked by transposed convolution exist in each channel of the decoder layers. The max-unpooling operation is partial inverse of the max-pooling, which is an upsampling process, transferring its input value to one of the locations in its output feature map and zeroing all the other locations. The non-zero location is derived from the max-pooling stage. The indices of max locations for each max-pooling operation in the encoder are stored and used in the corresponding max-unpooling operation as the non-zero location. In the max-unpooling, the output feature map is generated assigning the saved index position with its input and zeroing all non-maximal values. The kernel size of all convolution filters is chosen $3$ with zero padding of size $1$. In all pooling layers, the kernel size and the stride are considered $2$. The $29\times1$ expression or $199\times1$ shape parameters are fed to the proposed network. After passing four layers of encoder, the final encoded parameters are a $1\times64$ or $12\times64$ vector in the expression and shape networks, respectively. Then this representation is fed to the decoder in order to reconstruct both the clean expression and the shape parameters.

\subsection{Dataset}
For network training, a dataset containing pairs of correct and noisy 3DMM parameters is required. The best way to insure that the 3DMM parameters considered as clean ones would properly represent a possible 3D face structure, is making use of fitted 3DMM parameters to real face images as the ground truth data which are available in the datasets. Two different datasets are used in order to prepare the required training data: 300W-3D \cite{zhu2016face} and AFLW2000-3D \cite{zhu2016face}.

\textbf{300W-3D:} 300W \cite{sagonas2013300} contains images from multiple databases with their corresponding standardized $68$-point landmarks. 300W-3D consists of $3,837$ images of 300W with their fitted 3DMM parameters.  Zhu \emph{et. al.} have employed the Multi Features Framework (MFF) \cite{romdhani2005estimating} to provide the fitted 3DMM parameters of 300W samples \cite{zhu2016face}.

\textbf{AFLW2000-3D:}
AFLW \cite{koestinger2011annotated} contains $21,080$ face images in large poses with their annotated $21$-point visible landmarks. In \cite{zhu2016face}, a reconstruction algorithm is applied to the first $2000$ samples of AFLW resulting in AFLW2000-3D database, which contains the fitted 3DMM parameters and their corresponding $68$-point landmarks.

Putting all samples of the above two datasets together form a large dataset containing $5,837$ fitted 3DMM parameters to the real face images. In order to train the shape and expression networks, two different datasets are needed, one with noisy shapes and the other with noisy expressions. The mentioned fitted shape and expression parameters are the clean and noiseless parameters and are considered as ground truth data which would be compared to the network outputs. Noisy versions of the parameters are also required as the inputs to the network.

A simple yet effective procedure is employed in generating noisy versions of the parameters. In order to randomly corrupt the expression parameters, first a number $n$ is selected randomly between $1$ and $29$. $n$ is defined as the number of parameters to be corrupted with noise. Then, $n$ parameters are randomly selected from the expression parameters. Afterwards, different Gaussian noise signals with standard deviation denoted as $\sigma_{exp}$ are added to each of the selected parameters. The same process is repeated $50$ times for each of the samples of the dataset in order to prepare $50$ corrupted versions of each of the samples. Finally a dataset, containing $291,850$ pairs of clean and noisy 3DMM expression parameters is obtained. $261,850$ samples are used as training data and about $10\%$ or $30,000$ samples are considered as the test data.

\begin{figure}[!t]
	\centering
	\includegraphics[width=3.5in]{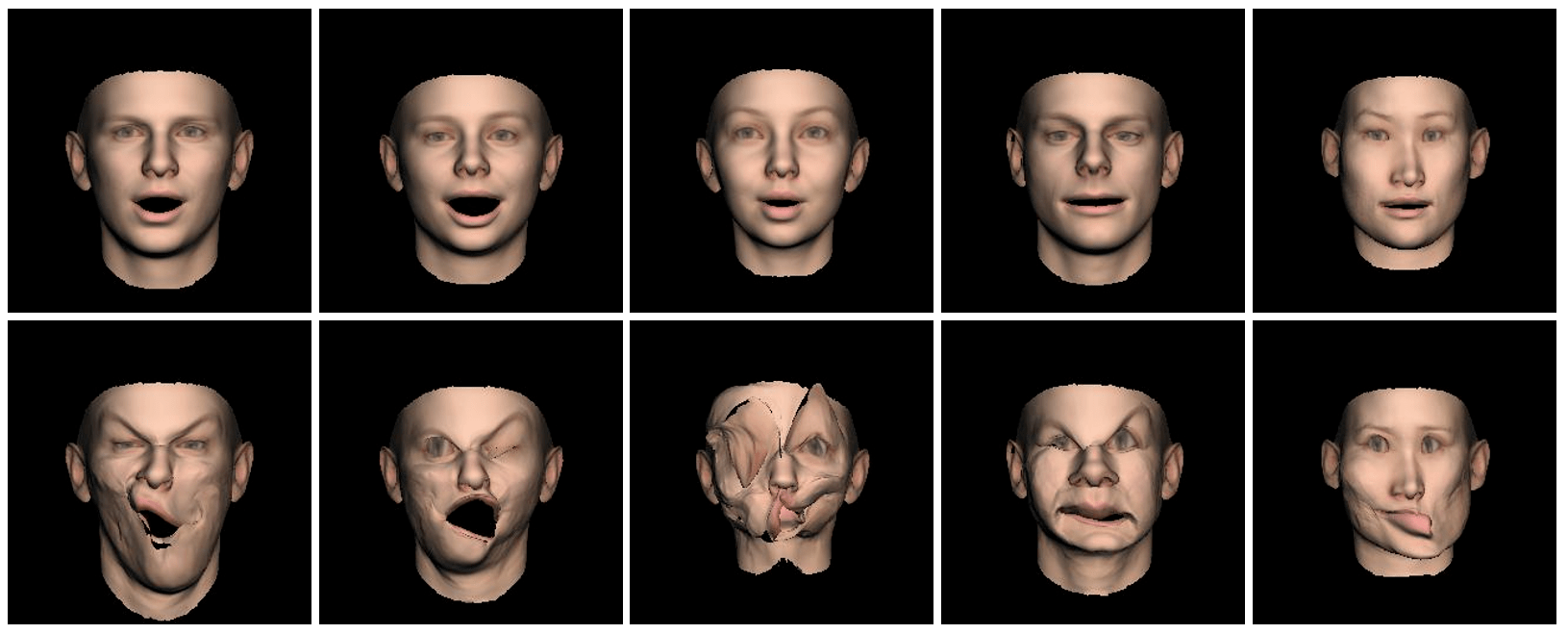}
	\caption{Some samples of the prepared expression dataset. Top row: faces with ground truth expression parameters. Bottom row: faces with noisy expression parameters.}\label{Fig2}
\end{figure}

\begin{figure}
	\centering
	\includegraphics[width=3.5in]{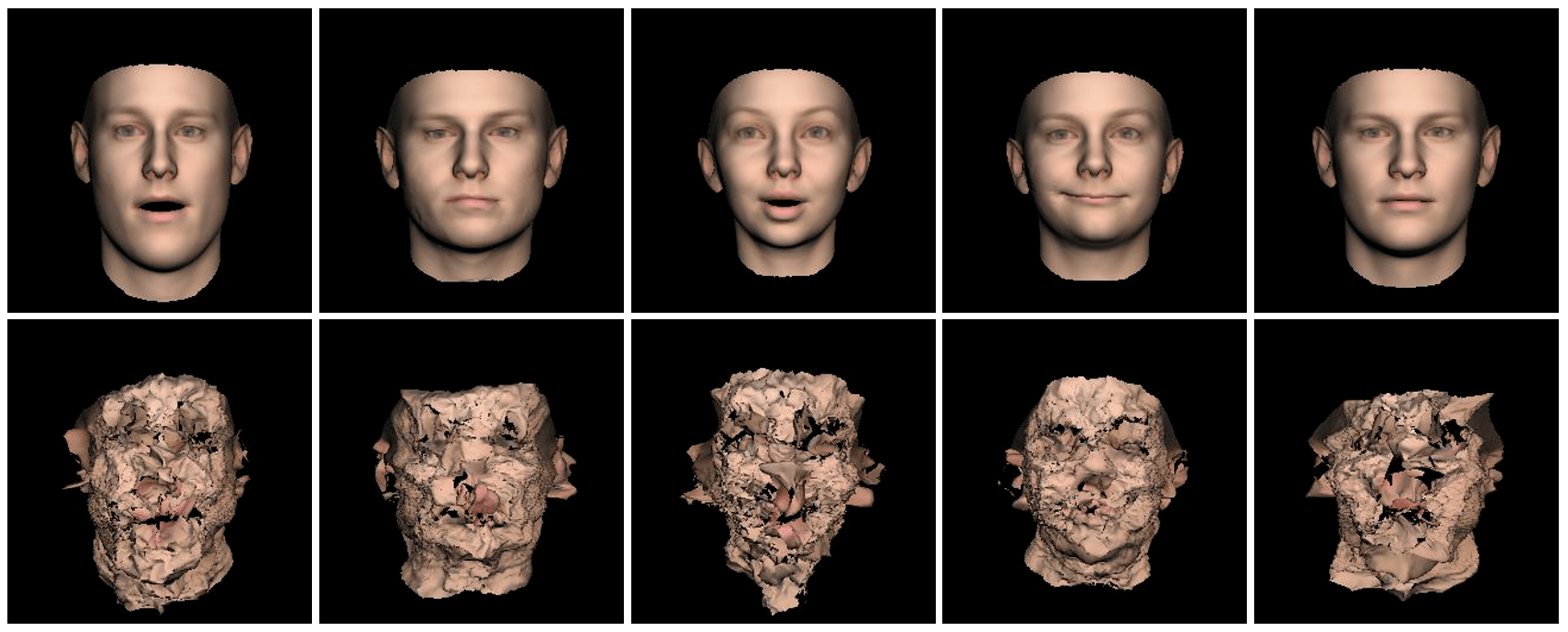}
	\caption{Some samples of the prepared shape dataset. Top row: faces with ground truth shape parameters. Bottom row: faces with noisy shape parameters.}\label{Fig6}
\end{figure}

The same procedure is done in the case of shape parameters. $\sigma_{shape}$ denotes the standard deviation of the added noise. The number of prepared noisy versions of each sample is $100$. Finally a dataset, containing $583,700$ pairs of clean and noisy 3DMM shape parameters is prepared. $523,700$ pairs are used as training data and the other $60,000$ samples are used as the test data. Some examples of the prepared shape and expression samples are shown in Fig. \ref{Fig2} and Fig. \ref{Fig6}, respectively.

This noisy dataset generation fashion is chosen because it leads to satisfying results in practice. The variance of the noise added to the training dataset completely depends on noise values expected to be removed by the network. As the interval of choosing 3DMM parameters in synthetic dataset generation scenario becomes larger, the generated faces would be more corrupted, hence elimination of larger noises would be required. In the cases where the noise variance added to training data is too small, the network would not be able to recognize and eliminate the noise signal.

\subsection{Training}
In the training procedure, the noisy 3DMM parameters are fed to the network, and then the network outputs are compared to the ground truth parameters. The mean squared error (MSE) between the network outputs and clean parameters is used as the loss function. The networks are trained for 10 epochs over the training data. The learning rate is set to 0.001 and the batch size is considered 128. Adam optimization algorithm \cite{Kingma2014} is employed for training the networks. As mentioned, shape parameters have very large values; thus to enable the network to learn from this data, the whole shape training set is divided by $10^5$. The trained network is evaluated in the next section. Training and evaluation were conducted on a desktop PC hardware with an NVIDIA GeForce GTX 1070 GPU and an Intel Core i7-4770K @ 3.50GHz CPU.

\section{Experiments} \label{sec:exp}
To verify the effectiveness of the proposed method, the experimental results would be presented in three subsections: qualitative and quantitative results on the test data, qualitative results on the corrupted uniformly generated synthetic dataset, and comparison between diversity of the synthetic datasets generated by the proposed method and the common traditional methods. In the third subsection, experiments show that the proposed method surpasses common methods in terms of diversity in shapes and expressions by a large margin.

\begin{figure}[!t]
	\centering
	\includegraphics[width=3in]{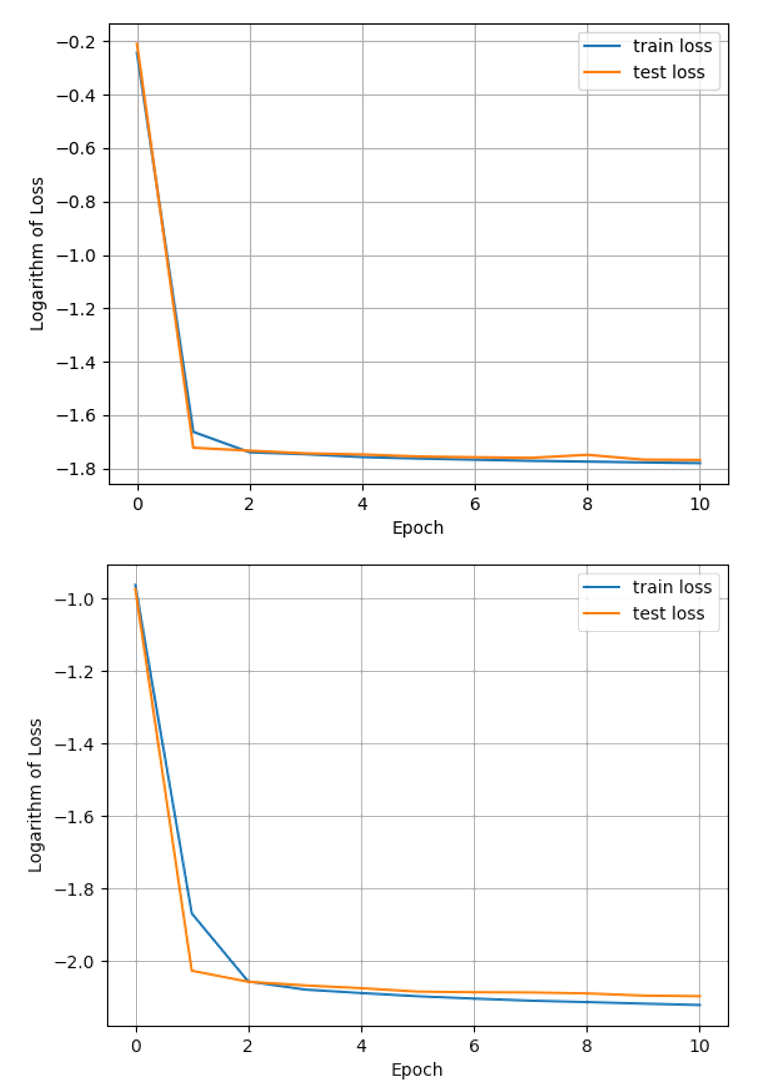}
	\caption{Logarithm of the training and testing loss over epochs. Learning curves of the shape (top) and expression (bottom) networks.}\label{Fig9}
\end{figure}

First of all, in order to generate training data, $\sigma_{shape}$ and $\sigma_{exp}$ must be obtained empirically. As expressed before, the amount of these parameters depends on intervals of uniform distribution in synthetic data construction. Here the network is expected to denoise synthetic data with the shape and expression parameters drawn from uniform distribution within intervals 10 and 15 times of their corresponding eigenvalues, respectively. The appropriate values for $\sigma_{shape}$ and $\sigma_{exp}$ are experimentally achieved $500,000$ and $2$ for this setup. The training and testing loss over learning epochs are plotted in Fig. \ref{Fig9}.

\begin{figure}[!t]
	\centering
	\includegraphics[width=3in]{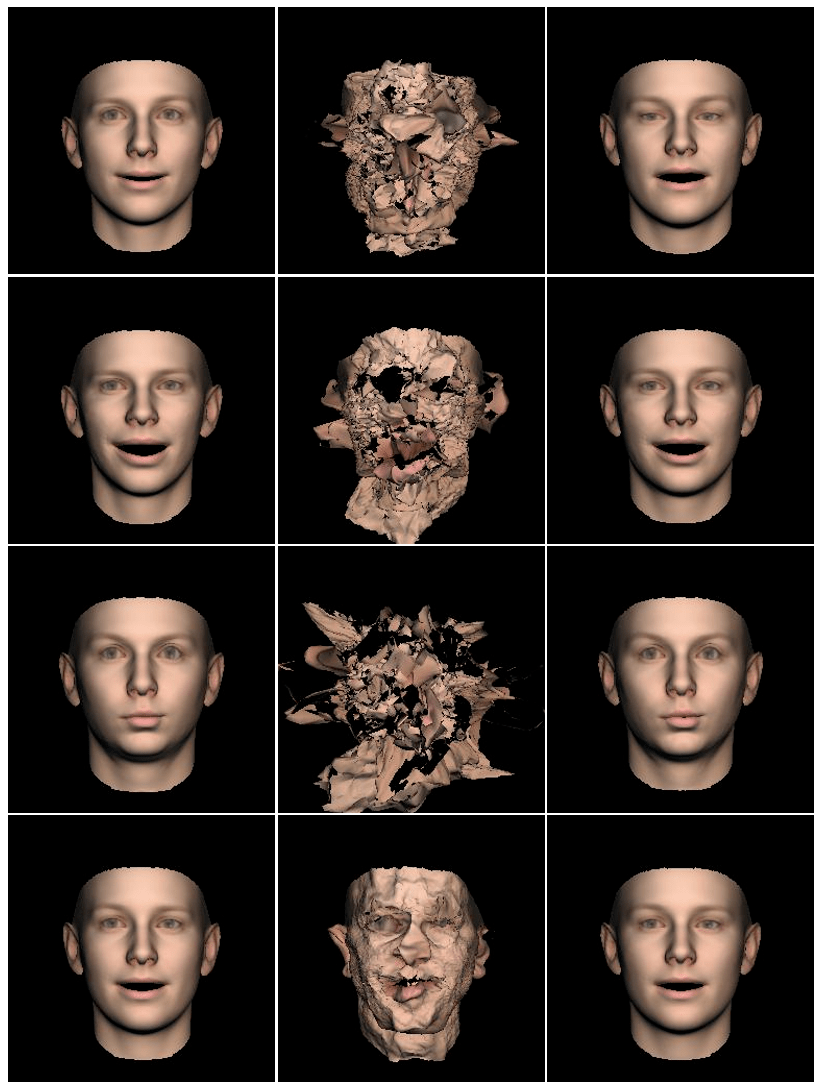}
	\caption{The results of the shape network on test data. Faces with ground truth shape parameters are in the left column, the second column contains the faces with noisy shape parameter as input to the network, and the third column shows the faces with network output as their shape parameters.}\label{Fig8}
\end{figure}

\begin{figure}[!t]
	\centering
	\includegraphics[width=3in]{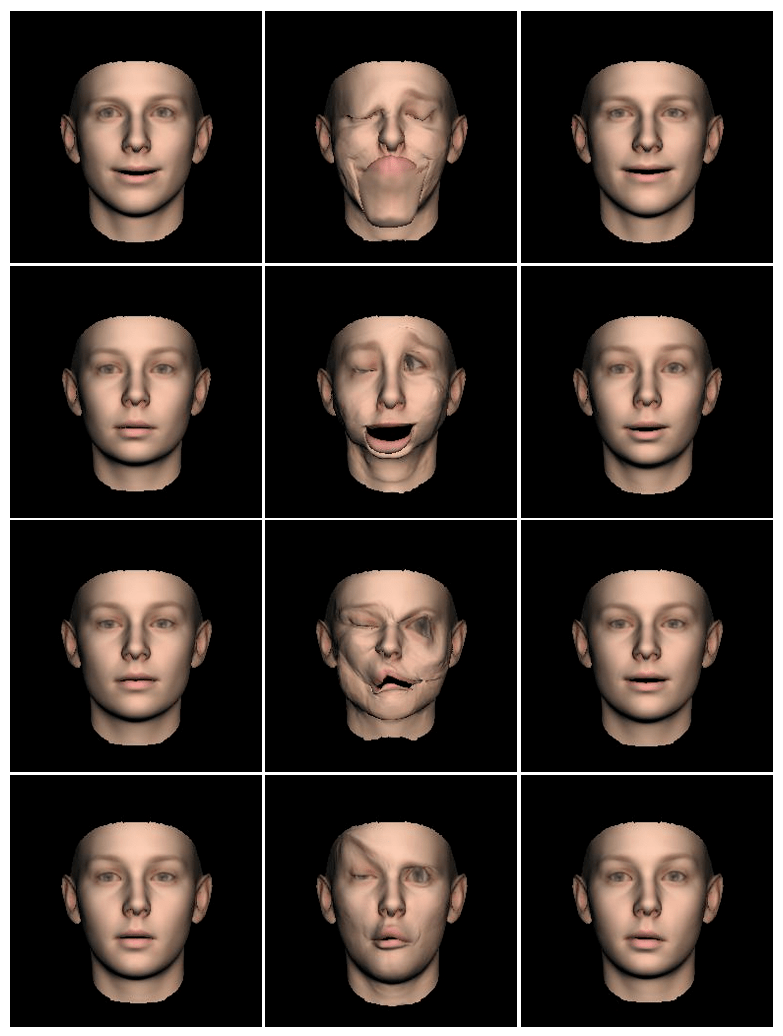}
	\caption{The results of the expression network on test data. Faces with ground truth expression parameters are in the left column, the second column contains the faces with noisy expression parameter as input to the network, and the third column shows the faces with network output as their expression parameters.}\label{Fig4}
\end{figure}

\begin{figure}[!t]
	\centering
	\includegraphics[width=3in]{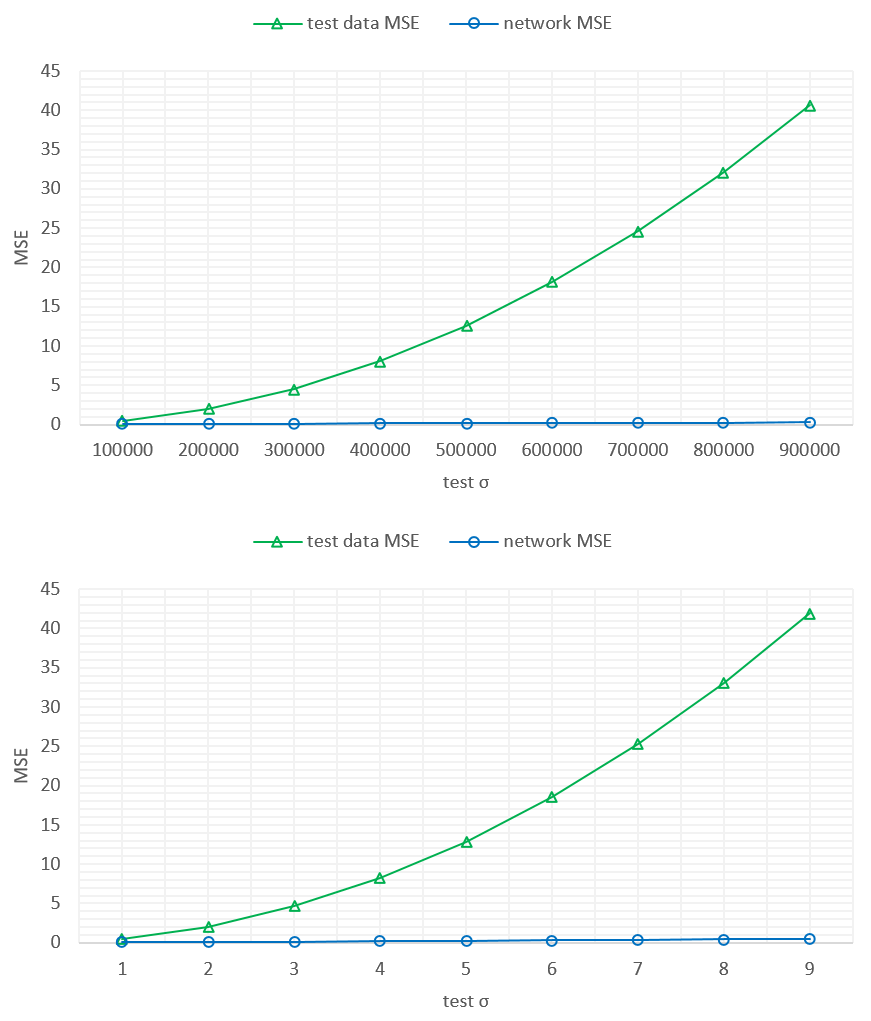}
	\caption{MSE of the test data and the network output over noise variance value of the test data for shape (top) and expression (bottom) networks.}\label{Fig10}
\end{figure}

\subsection{Result on Test Data}
To analyze the network accuracy in denoising test faces, the prepared test data for shape and expression are fed to their corresponding networks and the outputs of the networks are compared to the ground truth values. Mean squared error criterion is used as a performance measure to better examine the accuracy of the network output. This yields to MSE $0.12$ for the expression network and $0.17$ for the shape network. It is worth mentioning that the noisy and clean shape parameters are divided by $1e5$. The MSE of noisy test data fed to the network is $2.07$ for the expression and $12.57$ for the shape parameters. This sharp decline in MSE from $2.07$ to $0.12$ for expression, and from $12.57$ to $0.17$ for shape parameters, implies that the networks are able to successfully recover facial shape and expression parameters from noisy ones. To examine the performance of the proposed network on the test data with noise variance different than the noise variance of the training data, an experiment is conducted and its results are shown in Fig. \ref{Fig10}. In this figure the x-axis shows the variance of the test data. We note that the shape (top) and expression (bottom) networks are trained with the noise variance of 500,000 and 2, respectively. Increasing $\sigma$ in the test data generation procedure leads to a parabolic increase of the test data MSE. However, the proposed networks are able to reduce the noise MSE to around zero, and still remain roughly constant in a wide range of noise variance values. This property induces completely clean faces generated by the network even when extremely noisy faces are fed to it. The qualitative results of evaluation on test data are shown in Fig. \ref{Fig8} and Fig. \ref{Fig4} for the shape and expression networks, respectively. It can be observed that both networks are able to reconstruct noiseless faces with completely clean shapes and expressions. Moreover, the original shape and expression are preserved almost as they were before adding noise.

\begin{figure}[!t]
	\centering
	\includegraphics[width=3.5in]{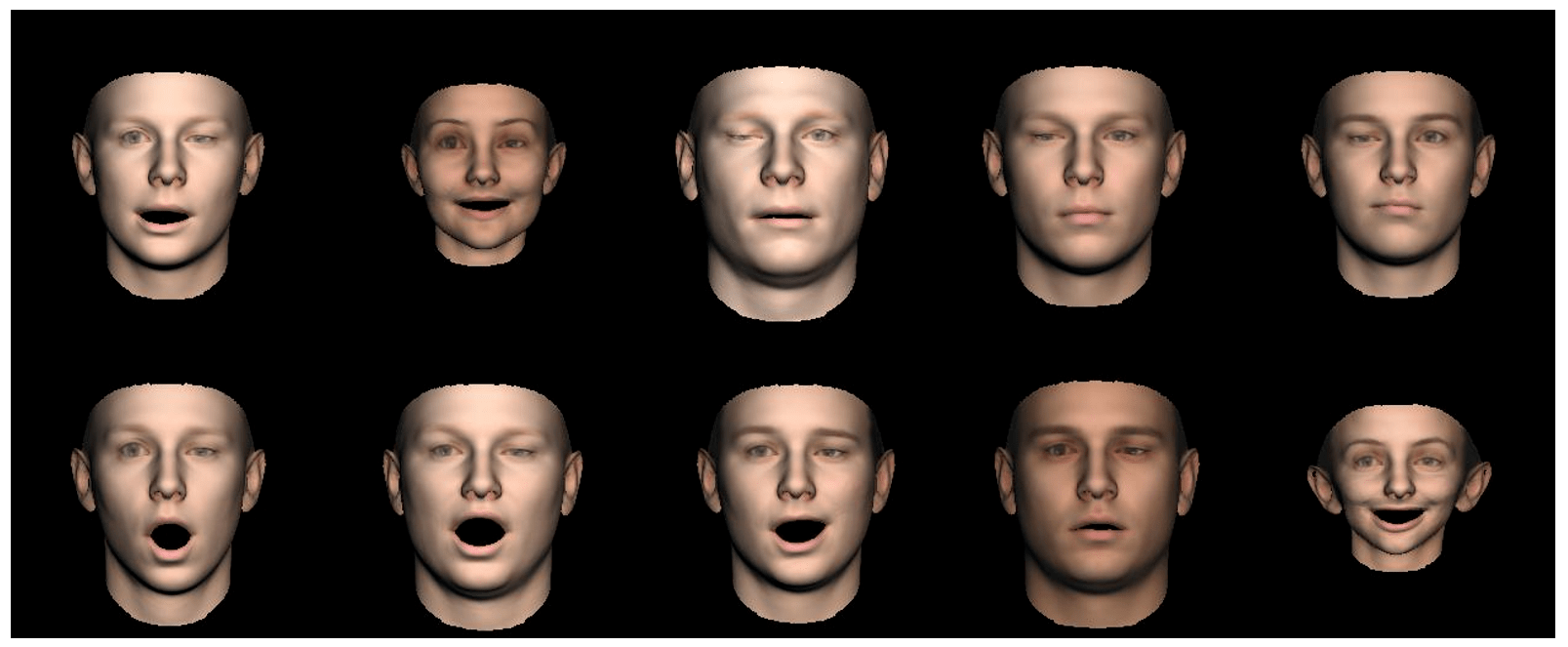}
	\caption{The only failures of the network in denoising 3D faces among 500 faces. The bottom right image is accounted as the shape network failure and in the other images the facial expression is corrupted.}\label{fig:Fig12}
\end{figure}

\begin{figure}[!b]
	\centering
	\includegraphics[width=3in]{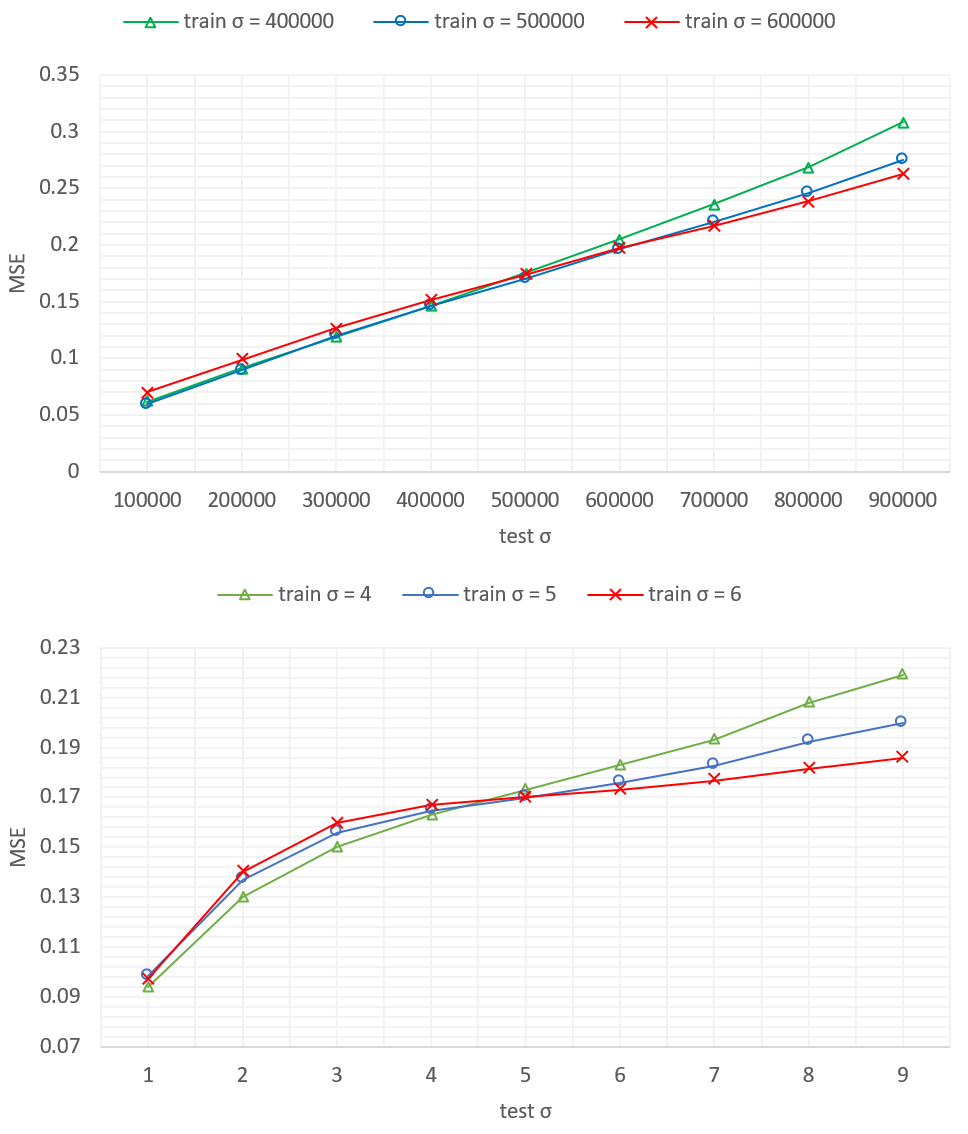}
	\caption{Comparison the MSE of the networks trained on datasets with three different noise levels against $\sigma$ of the test noise for shape (top) and expression (bottom) networks.}\label{Fig11}
\end{figure}

In order to assess the networks performance in the situation where the variance of the noise added to the test data is not equal to the noise variance used in training data, further analysis of noisy data variance would be conducted in this part. We train different networks on three different values of the noise variance used in the training data generation procedure. The test data are also generated with nine different values of noise variance. Then, the MSE of the networks output on each set of test data is measured. The results of this experiment are demonstrated in Fig. \ref{Fig11}. In this figures, the x-axis shows the variance of the noise added to the test data. As observed, all curves are ascending, implying that increasing the test data noise level would reduce the performance; however, the performance distance is insignificant if compared to the MSE of input data as illustrated in Fig. \ref{Fig10}. Besides, the networks trained on noisy data with larger noise levels have the best performance on data with high noise level and vice versa. But for values of noise level less than a certain value, all networks lead to almost the same output MSE, i.e. eliminating low level of noise would be a simple task for all networks.

\subsection{Result on Synthetic Dataset}
\begin{figure}[!t]
	\centering
	\includegraphics[width=3.5in]{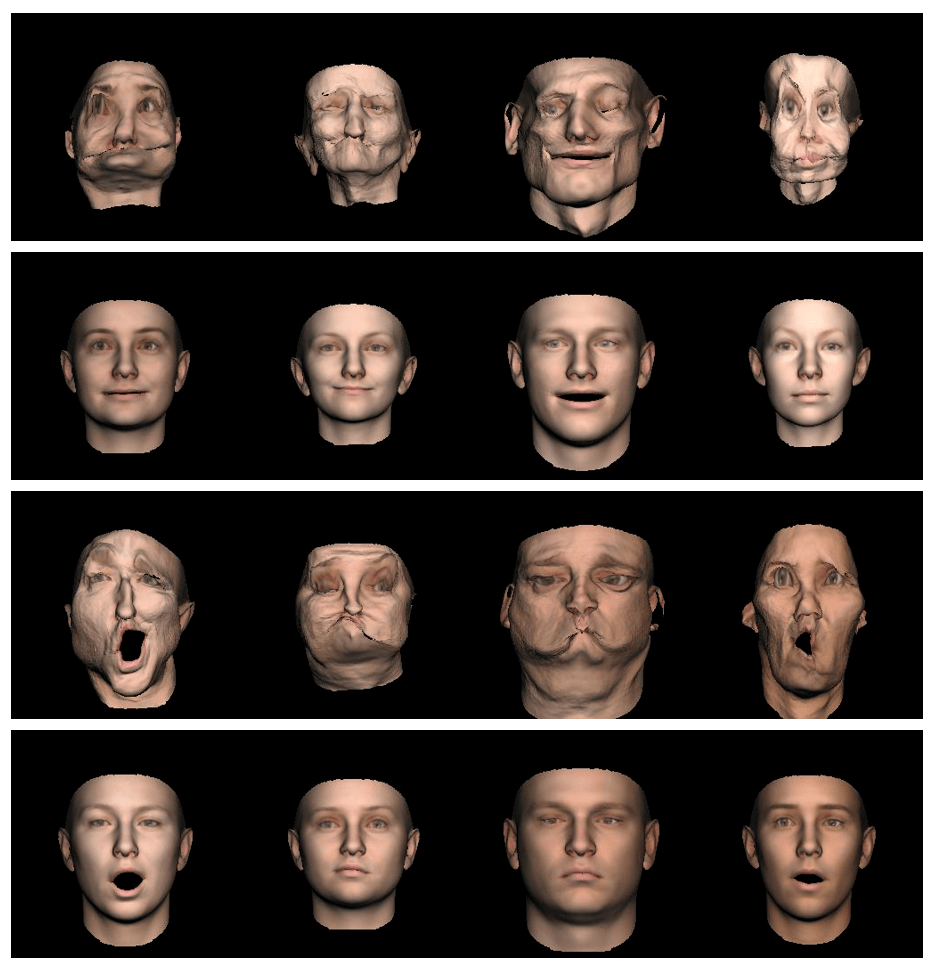}
	\caption{The results of the proposed method on a synthetic dataset with uniformly distributed 3DMM parameters. The interval of the uniform distribution is considered 10 times eigenvalues for the shape and 15 times eigenvalues for the expression parameters. The first and third rows contain generated synthetic faces, and the second and fourth rows contain the faces generated by feeding each of the corrupted faces to the proposed networks.}\label{Fig5}
\end{figure}

Here, we are willing to show the performance of the proposed networks in generating synthetic face as the main purpose of our study. In the following, qualitative results on noisy synthetic dataset is provided. Experiments show that the proposed method obtains high quality synthetic face generation. Synthetic dataset images are generated by drawing random expression and shape parameters from uniform distribution in intervals $15$ and $10$ times their corresponding eigenvalues around zero. This is the same procedure as faces generated in Fig. \ref{Fig1}). These noisy expression and shape parameters are fed as the inputs to each of the networks and the outputs are examined qualitatively (see Fig. \ref{Fig5}). The networks has generated clean and noiseless outputs. Thereby this method enables us to create a synthetic dataset with high diversity in expressions and shapes without any corruption at the same time.

To better evaluate the reliability of the network outputs which will be considered as synthetic face data, a quality metric is introduced. The metric is defined as the percentage of acceptable outputs as a real face in the generated synthetic dataset. The percentage is computed feeding 500 noisy face shape and expression parameters to their corresponding networks and examining the generated faces. The experiment is done using 30 subjects, evaluating the generated faces. A sample would be considered as \textit{non-possible} face, if at least one subject rejects the face. This leads to $98.2\%$ reliability for the expression network and $99.8\%$ for the shape network. Fig. \ref{fig:Fig12} represents all of the outputs rejected to be real faces in this experiment. The generated expressions are not accepted only because of a minor asymmetry in the eyes. It is possible to increase the reliability of the networks outputs, in exchange of their diversity by reducing range of uniform distribution of the input parameters. For instance, if the expression parameters are drawn from uniform distribution within 5 times their eigenvalues around zero, the expression network generates real outputs for all 500 faces.

\begin{figure}[!t]
	\centering
	\includegraphics[width=3.5in]{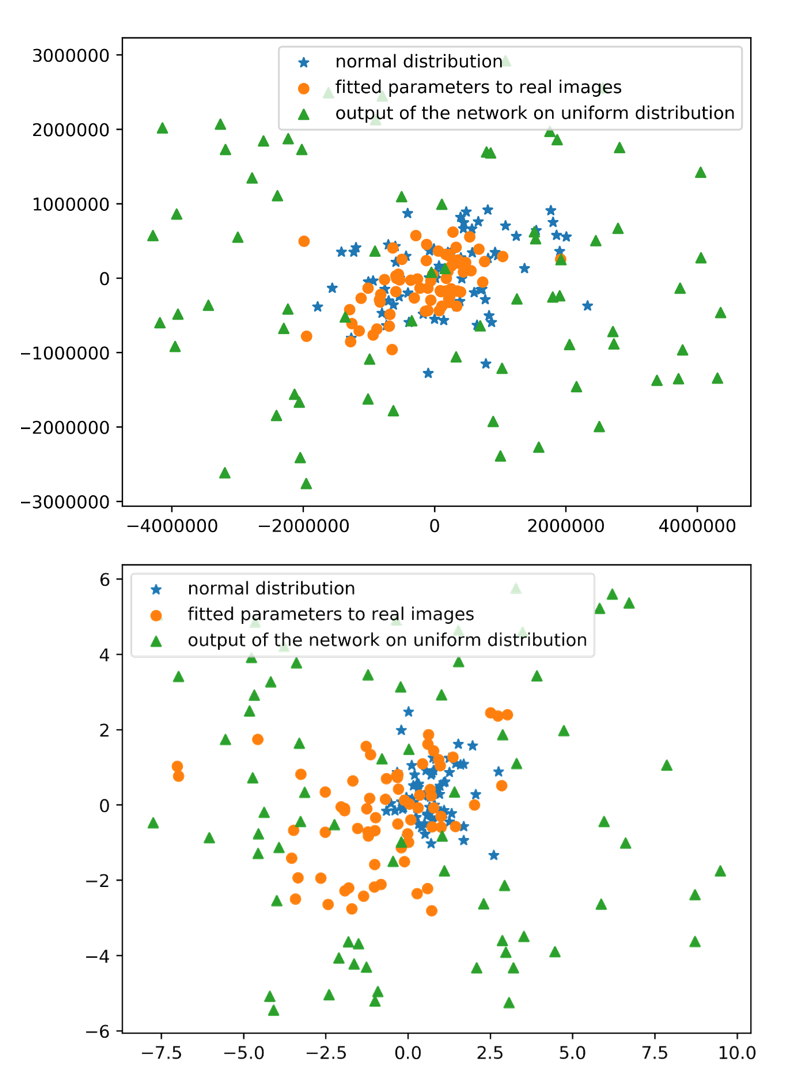}
	\caption{Scatter diagrams comparing diversity of three datasets in terms of shape (top) and expression (bottom).}\label{Fig7}
\end{figure}

\subsection{Scatter Diagrams}
This experiment proves that the proposed method for generating synthetic dataset leads to increased scattering and diversity of the generated faces compared to the previous methods. In order to plot scatter diagram, PCA dimensionality reduction is applied to the shape and expression parameters separately. The same work is repeated for three different datasets: 1) the dataset generated by feeding randomly generated parameters from a uniform distribution to the networks (the proposed method), 2) the dataset generated by choosing parameters from a normal distribution, and 3) the fitted 3DMM parameters to the real images. The comparison between diversity of these three datasets is shown as the scatter diagram plotted for 70 samples in Fig. \ref{Fig7}. As can be seen, the scatter diagram of the generated dataset by using the proposed method is higher than the other two datasets. Thus, it can be demonstrated that the proposed method for synthetic face generation outperforms the existing methods in terms of diversity of shapes and expressions. It should be noted that the amount of the scattering is completely adjustable by changing the uniform distribution interval used in the proposed procedure.

\begin{table}[!t]
	\renewcommand\arraystretch{1.5}
	\centering
	\caption{Quantitative comparison between the diversity of three datasets.}
	\begin{tabular}{ccccl}
		\cline{1-4}
		\multicolumn{1}{|c|}{\textbf{Parameters}} & \multicolumn{1}{c|}{\textbf{Our dataset}} & \multicolumn{1}{c|}{\textbf{Normal dataset}} & \multicolumn{1}{c|}{\textbf{Realistic dataset}} &  \\ \cline{1-4}
		\multicolumn{1}{|c|}{\textbf{shape}} & \multicolumn{1}{c|}{\equation{\textbf{15.54e12}}} & \multicolumn{1}{c|}{$1.87e12$} & \multicolumn{1}{c|}{$0.86e12$} &
		\\
		\multicolumn{1}{|c|}{\textbf{expression}} & \multicolumn{1}{c|}{\equation{\textbf{47.58}}} & \multicolumn{1}{c|}{$2.48$} & \multicolumn{1}{c|}{$7.08$} &  \\ \cline{1-4}
		\multicolumn{1}{l}{} & \multicolumn{1}{l}{} & \multicolumn{1}{l}{} & \multicolumn{1}{l}{} &
	\end{tabular}
	\label{scatter}
\end{table}

A scattering criterion is also used for a quantitative comparison. The trace of the covariance matrix is considered as the scattering criterion. The value of this criterion for 2000 samples of each of the three datasets is reported in Table \ref{scatter} which confirms the scatter diagram results.

\section{Conclusion} \label{sec:con}
3D face reconstruction using 2D images is a challenging task where deep learning techniques achieved promising results. Synthetic dataset generation is a way to prepare the training data sets for such techniques. In this paper, we proposed an autoencoder network which learns the human face manifold to generate synthetic faces. Using manifold, makes the network capable to deal with the \textit{non-possible} faces. We showed that the common existing methods for synthetic face generation reduce diversity in order to have non-corrupted faces. The proposed network produces completely possible faces without sacrificing the diversity. In other words, it generates highly diverse dataset without any \textit{non-possible} faces. Experiments show that the diversity of the generated faces could be improved up to 8 and 19 times in terms of the shape and expression, respectively, in comparison to the the existing methods and datasets. Also it could be further improved by adding more noises to the training datasets. Experiments also confirm that the trained network is robust against high MSE values of noise and can denoise highly corrupted faces. The high reliability in the results ($98.2\%$ in the expression and $99.8\%$ in the shape of the generated faces) insures that the proposed network could be placed at the top of any 3D reconstruction network to improve the output quality.


%

\ifCLASSOPTIONcaptionsoff
  \newpage
\fi



\bibliographystyle{IEEEtran}
\bibliography{ref}
%

%





\end{document}